\documentclass{article}

\usepackage{microtype}
\usepackage{graphicx}
\usepackage{subcaption}
\usepackage{booktabs} % for professional tables

\usepackage{hyperref}

\usepackage[preprint]{icml2026}

\usepackage{amsmath}
\usepackage{amssymb}
\usepackage{mathtools}
\usepackage{amsthm}
\usepackage{booktabs}
\usepackage{multirow}
\usepackage{tabularx}  % 用于自适应宽度表格
\usepackage{graphicx}  % 用于\resizebox（如果使用第一个版本）
\usepackage[table]{xcolor}  % 支持表格颜色
\usepackage{colortbl}        % 表格颜色支持
\usepackage[normalem]{ulem}

\usepackage{xspace}
\newcommand{\method}{\texttt{SOAR}\xspace}
\newcommand{\methodfull}{\textbf{S}earch \textbf{O}r \textbf{A}ccele\textbf{R}ate (\method)\xspace}

\usepackage[capitalize,noabbrev]{cleveref}

\theoremstyle{plain}
\newtheorem{theorem}{Theorem}[section]

\theoremstyle{definition}
\newtheorem{definition}[theorem]{Definition}

\theoremstyle{remark}

\usepackage[textsize=tiny]{todonotes}

\icmltitlerunning{Confidence-Switched Position Beam Search for DLM}

\begin{document}

\twocolumn[
  \icmltitle{Search or Accelerate: Confidence-Switched Position Beam Search for Diffusion Language Models}
  \icmlsetsymbol{equal}{*}

  \begin{icmlauthorlist}
    \icmlauthor{Mingyu Cao}{a}
    \icmlauthor{Alvaro H.C. Correia}{b}
    \icmlauthor{Christos Louizos}{b}
    \icmlauthor{Shiwei Liu}{c,d,e}$^\dagger$,
    \icmlauthor{Lu Yin}{a}$^\dagger$,

  \end{icmlauthorlist}

  \icmlaffiliation{a}{University of Surrey}
  \icmlaffiliation{b}{Qualcomm AI Research. Qualcomm AI Research is an initiative of Qualcomm Technologies, Inc.}
  \icmlaffiliation{c}{ELLIS Institute Tübingen}
  \icmlaffiliation{d}{Max Planck Institute for Intelligent Systems}
  \icmlaffiliation{e}{Tübingen AI Center}

  \icmlcorrespondingauthor{Lu Yin}{l.yin@surrey.ac.uk}
  \icmlcorrespondingauthor{Shiwei Liu}{sliu@tue.ellis.eu}

  % You may provide any keywords that you find helpful for describing your
  % paper; these are used to populate the "keywords" metadata in the PDF but

  \vskip 0.3in
]

% this must go after the closing bracket ] following \twocolumn[ ...

% This command actually creates the footnote in the first column listing the
% affiliations and the copyright notice. The command takes one argument, which
% is text to display at the start of the footnote. The \icmlEqualContribution
% command is standard text for equal contribution. Remove it (just {}) if you
% do not need this facility.

% Use ONE of the following lines. DO NOT remove the command.
% If you have no special notice, KEEP empty braces:
\printAffiliationsAndNotice{}  % no special notice (required even if empty)
% Or, if applicable, use the standard equal contribution text:
% \printAffiliationsAndNotice{\icmlEqualContribution}

% \begin{abstract}

% Diffusion Language Models (DLMs) have emerged as strong competitors to autoregressive models. The standard decoding process for DLMs greedily selects the mask token with the highest confidence at each step for unmasking, which is limiting. While beam search has long improved autoregressive decoding by expanding the search space, its adaptation to DLMs is challenging due to their non-sequential, position-parallel nature. To bridge this gap, we adapt beam search to DLMs with a key difference: the search operates across the position dimension rather than sequentially over tokens. To realize this, we introduce \methodfull, a confidence-switched position beam search procedure that dynamically adjusts beam width based on model confidence, enabling both broader search and faster parallel decoding. When the model is uncertain, \method explores multiple unmasking paths and widens the beam, thereby improving global confidence. When highly confident, it decodes multiple tokens in parallel and narrows the beam to accelerate inference. Experimental results on mathematics and coding benchmarks demonstrate that, in most cases, \method maintains faster decoding speed than standard methods while achieving higher reasoning quality, providing a practical, training-free approach to navigating the quality--speed trade-off in DLM inference.

% \end{abstract}

\begin{abstract}
Diffusion Language Models (DLMs) generate text by iteratively denoising a masked sequence, repeatedly deciding \emph{which positions} to commit at each step. Standard decoding follows a greedy rule, unmask the most confident positions, yet this local choice can lock the model into a suboptimal unmasking order, especially on reasoning-heavy prompts. We present \methodfull, a \textbf{training-free} decoding algorithm that adapts its behavior to the model’s uncertainty. When confidence is low, \method briefly widens the search over alternative unmasking decisions to avoid premature commitments; when confidence is high, it collapses the search and decodes many positions in parallel to reduce the number of denoising iterations. Across mathematical reasoning and code generation benchmarks (GSM8K, MBPP, HumanEval) on \textsc{Dream}-7B and \textsc{LLaDA}-8B, \method improves generation quality while maintaining competitive inference speed, offering a practical way to balance quality and efficiency in DLM decoding. Our Code is available at 
\url{https://github.com/duterscmy/SOAR}
\end{abstract}

\section{Introduction}
\label{sec:introduction}
\begin{figure}[t]
    \centering
    \includegraphics[width=0.85\linewidth]{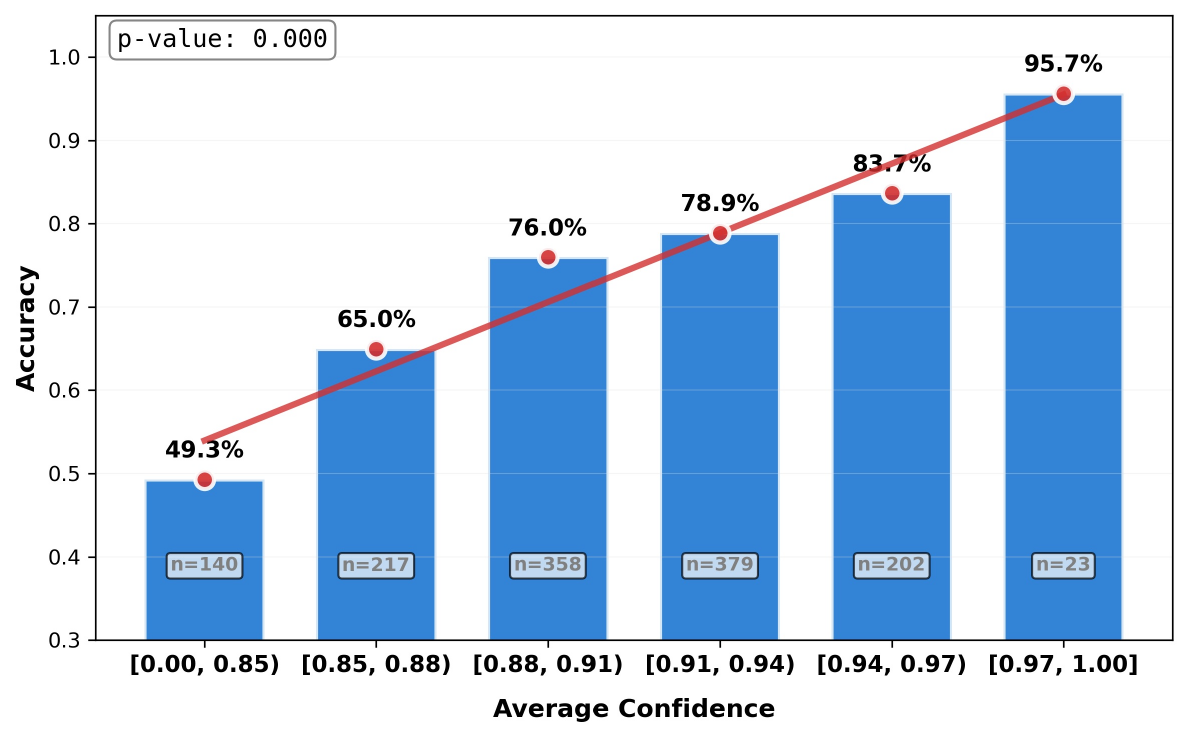}
    \vskip 0.1in
    \caption{Accuracy and average confidence on GSM8K with \textsc{Dream}-7B-Base. All questions are divided into 6 bins by their average decoding confidence, with $n$ indicating the sample size in each bin. The red solid line represents the trend line.}
    \vskip 0.1in
    \label{fig:confidence_with_acc}
\end{figure}

Diffusion Language Models (DLMs) \cite{ye2025dream7bdiffusionlarge,nie2025largelanguagediffusionmodels,zhu2025llada15variancereducedpreference} have recently emerged as a promising alternative to autoregressive (AR) generation \cite{touvron2023llamaopenefficientfoundation,yang2025qwen3technicalreport}. Unlike AR models that decode tokens sequentially in a fixed left-to-right order, DLMs iteratively refine a partially masked sequence and can decode multiple token positions in parallel. In practice, most mask-based DLM decoders follow a simple greedy rule: at each denoising step, they unmask the positions with the highest prediction confidence and keep the remaining positions as \texttt{[MASK]}.

This greedy unmasking rule is attractive for efficiency, but it also makes a strong local decision about \emph{which positions to commit next}. Importantly, for DLMs this unmasking order is not fixed (unlike left-to-right AR decoding), and different unmasking schedules can lead to different final generations. This motivates our central question: \textbf{Can we improve DLM decoding by explicitly exploring alternative unmasking orders beyond greedy selection?}

A natural signal for guiding such exploration is the model's prediction confidence, which has been leveraged in prior work for adaptive-parallel decoding \cite{chen2025dparallellearnableparalleldecoding}. In our setting, we observe the same qualitative pattern: on GSM8K \cite{cobbe2021trainingverifierssolvemath}, examples decoded with higher average token confidence tend to be more accurate (Figure~\ref{fig:confidence_with_acc}). This suggests that searching for decoding trajectories that maintain higher confidence may improve downstream quality, especially on reasoning-heavy prompts.

Inspired by the success of beam search in AR decoding \cite{Freitag_2017}, we adapt this paradigm to DLMs. A key difference is that, while AR beam search expands candidates primarily in the token/vocabulary space, DLM decoding admits a distinct search dimension: the \emph{position space}, i.e., the order in which masked positions are unmasked. We find that widening search in this position space consistently improves output quality across benchmarks, but it also increases computation roughly linearly with beam width. This leads to a practical challenge: \textbf{How can we retain the gains from position-space search without incurring prohibitive inference latency?}

To address this, we propose \method, a \textbf{training-free} confidence-switched position beam search procedure that dynamically balances exploration and speed based on the model's confidence at each step. When the model is uncertain, \method widens the beam to explore multiple unmasking choices and avoid premature commitments; when the model is confident, it decodes many high-confidence positions in parallel and collapses the beam to accelerate inference. This enables the model to spend compute when needed and decode quickly when possible, yielding a practical quality--speed trade-off for DLM inference.

Our contributions are summarized as follows:
\begin{itemize}
\item We introduce \textbf{P}osition \textbf{B}eam \textbf{S}earch (PBS), a training-free decoding method that explores alternative unmasking sequences in the position space and consistently improves output quality across multiple benchmarks.

\item We propose \method, a training-free adaptive inference algorithm that switches between parallel decoding and position-space search based on per-step confidence, effectively balancing speed and quality.

\item We demonstrate that \method integrates naturally with variable-length DLM decoding methods and multiple unmasking metrics, highlighting its flexibility and compatibility with advanced decoding strategies.
\end{itemize}

\section{Related Works}

\subsection{Diffusion Language Models}
Diffusion Language Models have emerged as a competitive alternative to autoregressive models for text generation \cite{nie2025largelanguagediffusionmodels,ye2025dream7bdiffusionlarge,gong2025scalingdiffusionlanguagemodels}. Unlike their autoregressive counterparts \cite{touvron2023llamaopenefficientfoundation, yang2025qwen3technicalreport} that generate tokens sequentially, DLMs define a forward process that gradually corrupts text with noise (or [MASK] tokens) and learn a reverse process to reconstruct the original text through iterative denoising.

The development of large-scale DLMs has progressed along several distinct architectural paths. Two prominent approaches have emerged based on their initialization strategies: methods that leverage pre-trained autoregressive models and those trained from scratch. DiffuLLaMA \cite{gong2025scalingdiffusionlanguagemodels} and \textsc{Dream} \cite{ye2025dream7bdiffusionlarge} follow a transfer learning paradigm, where DLMs are initialized from existing strong autoregressive models (LLaMA \cite{touvron2023llamaopenefficientfoundation} and Qwen \cite{yang2025qwen3technicalreport}, respectively). Their success demonstrates the viability of this approach, showing that DLMs can achieve competitive performance relative to strong autoregressive baselines. In contrast, \textsc{LLaDA} \cite{nie2025largelanguagediffusionmodels} represents a distinct path, being trained from scratch as a high-performing open-source DLM. Together, these works have established DLMs as a credible and high-performance architecture for text generation.The standard decoding process for these mask-based DLMs is iterative and parallel. Starting from a fully masked sequence, the model predicts all tokens simultaneously at each denoising step. A common strategy involves selecting and committing the tokens at the most confident positions, while replacing the remaining uncertain positions with \texttt{[MASK]} tokens for the next round of refinement. This "predict-all, then commit-a-subset" loop repeats until the entire sequence is unmasked, leveraging the model's bidirectional context at every step to refine predictions.

\subsection{Optimized Inference Methods for DLLMs}
The non-autoregressive, iterative nature of DLM decoding presents unique inference challenges, as the random access pattern across denoising steps invalidates the traditional KV-cache and necessitates pre-defining a maximum sequence length, often leading to computational redundancy. Thus, research has focused on specialized acceleration techniques. 

One primary direction involves designing specialized KV-cache mechanisms that either leverage the high similarity of hidden states across steps for approximate caching or restructure decoding into a block-autoregressive manner to enable state reuse from previous context \cite{wu2025fastdllmtrainingfreeaccelerationdiffusion,nguyentri2025attentionneedkvcache,huang2025masktokensprophetfinegrained,liu2025dllmcacheacceleratingdiffusionlarge,ma2025dkvcachecachediffusionlanguage}. Another critical direction aims to reduce the total number of steps through parallel and adaptive-length decoding. This includes confidence-based strategies that dynamically determine how many tokens to decode in parallel, as well as early-commit methods which halt decoding once the model's predictions stabilize, significantly speeding up inference \cite{wu2025fastdllmtrainingfreeaccelerationdiffusion, yang2025diffusionllmnativevariable, gao2025selfspeculativedecodingdiffusion, li2025diffusionlanguagemodelsknow, li2025fixedtrainingfreevariablelengthdenoising}. In addition, some works address the inherent inability of standard DLMs to revise committed tokens, exploring solutions ranging from training-free correctors to modifications of the diffusion process itself \cite{huang2025dontsettleearlyselfreflective,mounier2025reviewremaskrefiner3,zhu2025latentrefinementdecodingenhancing}. 
A concurrent work, Order-Token Search \cite{anonymous2025improving}, proposes to simultaneously explore the decoding space across both the position and token dimensions to achieve better results, but at the expense of inference speed. Our proposed method, \method, dynamically switches between search and parallel decoding modes based on the model's confidence during token decoding. This enables consistent improvements across multiple benchmarks without sacrificing decoding speed.

\section{Methodology}

\subsection{Preliminaries of DLM Inference}
Given a text sequence of length $L$, we denote $x = [x_1, x_2, \dots, x_L] \in \mathcal{V}^L$ where $\mathcal{V}$ is the vocabulary. In diffusion language models, the forward process gradually corrupts $x$ through $T$ steps:

\begin{equation}
q(x_t | x_{t-1}) = \begin{cases}
\mathcal{M}(x_{t-1}, \rho_t) & \text{if } t > 0 \\
\delta(x) & \text{if } t = 0
\end{cases}
\end{equation}
 
where $\mathcal{M}(\cdot, \rho_t)$ masks a proportion $\rho_t$ of tokens, and $\delta(\cdot)$ is the Dirac delta function. The reverse process iteratively reconstructs the original text by modeling the reverse transition as conditionally independent across positions in $x_{t-1}$ given the corrupted sequence $x_t$:

\begin{equation}
p_\theta(x_{t-1} | x_t) = \prod_{i=1}^L p_\theta(x_{t-1,i} | x_t),
\end{equation}

where $x_{t,i}$ denotes the $i$-th token at step $t$, and $p_\theta$ is parameterized by the DLM.

At inference time, starting from a fully masked sequence $x_T = [\text{[MASK]}]^L$, the model generates text through $T$ denoising iterations. At each step $t$, the model predicts a distribution over $\mathcal{V}$ for every position $i$:

\begin{equation}
\mathbf{p}_{t,i} = \text{softmax}(f_\theta(x_t)_i) \in \mathbb{R}^{|\mathcal{V}|}
\end{equation}
 
where $f_\theta$ is the DLM. The decoding strategy determines which tokens to decode at each step. 

Let $\mathcal{I}_t \subseteq \{1, \dots, L\}$ be the set of positions selected for unmasking at step $t$. The standard approach selects $k$ positions with the highest confidence scores:

\begin{equation}
\mathcal{I}_t = \operatorname{topk}\left(\{\max_{v \in \mathcal{V}} \mathbf{p}_{t,i}[v] : i \in M_t\}, k\right)
\end{equation}
where $M_t = \{i: x_{t,i} = \text{[MASK]}\}$ represents the set of masked positions. Then the updated sequence is:

\begin{equation}
x_{t-1,i} = \begin{cases}
\arg\max_{v \in \mathcal{V}} \mathbf{p}_{t,i}[v] & \text{if } i \in \mathcal{I}_t \\
\text{[MASK]} & \text{otherwise}
\end{cases}
\end{equation}

The key challenge is how to select $\mathcal{I}_t$ to maximize sequence quality while minimizing computational cost.

\subsection{Position Beam Search (PBS)}
\label{subsec:position-beam-search}

However, the overall confidence of sequences generated through greedy decoding may not be optimal. To obtain decoding paths with higher confidence, we introduce a training‑free method that adapts the classic beam‑search paradigm to a new dimension—the position space. The PBS is initialized with a single sequence consisting only of masked tokens and a score of zero: $\mathcal{B}_0 = \{(x_0, 0)\}$, where $x_0$ contains only \texttt{[MASK]} tokens.

Let $\mathcal{B}_t = \{(x^{(1)}_t, s^{(1)}_t), (x^{(2)}_t, s^{(2)}_t), \dots, (x^{(K)}_t, s^{(K)}_t)\}$ be defined as the beam of size $K$ at step $t$, where each $x^{(j)}_t$ is a candidate sequence and $s^{(j)}_t$ is its cumulative score, with $j$ indexing the candidate in the beam.

\subsubsection{Candidate Generation}
% For each candidate $(x^{(j)}_t, s^{(j)}_t)$ in the beam, new candidates are generated by selecting different sets of masked positions to unmask. Given $M^{(j)}_t$ as the masked positions for candidate $j$, all subsets $\mathcal{I} \subseteq M^{(j)}_t$ of size $k$ are considered:

% \begin{equation}
% \mathcal{C}^{(j)}_t = \Bigl\{ (x^{(j,\mathcal{I})}_{t-1}, s^{(j)}_{t} + \phi(\mathcal{I}, \mathbf{p}^{(j)}_t)) : \mathcal{I} \subseteq M^{(j)}_t, |\mathcal{I}| = k \Bigr\}
% \end{equation}

% where $\mathbf{p}^{(j)}_t$ represents the predicted distributions for candidate $j$, $x^{(j,\mathcal{I})}_{t-1}$ is the sequence obtained after unmasking positions in $\mathcal{I}$, and $\phi(\mathcal{I}, \mathbf{p})$ is defined as a scoring function:

% \begin{equation}
% \phi(\mathcal{I}, \mathbf{p}) = \frac{1}{|\mathcal{I}|} \sum_{i \in \mathcal{I}} \max_{v \in \mathcal{V}} \mathbf{p}_i[v]
% \end{equation}
% The average probability is used instead of the sum to avoid bias against sequences with more masked positions.

For each candidate $(x^{(j)}_t, s^{(j)}_t)$ in the beam, we generate new candidates by exploring different ways to unmask its currently masked tokens. Specifically, for each possible set of $n$ masked positions (denoted by $\mathcal{I}$) that we choose to unmask in this step, we obtain a new sequence $x^{(j)}_{t-1}$ and compute its cumulative score as:

\begin{equation}
s^{(j)}_{t} + \phi(\mathcal{I}, \mathbf{p}^{(j)}_t),
\end{equation}

where $\mathbf{p}^{(j)}_t$ represents the predicted token distributions for candidate $j$, and $\phi(\mathcal{I}, \mathbf{p})$ is the average confidence score of the selected positions:

\begin{equation}
\phi(\mathcal{I}, \mathbf{p}) = \frac{1}{|\mathcal{I}|} \sum_{i \in \mathcal{I}} \max_{v \in \mathcal{V}} \mathbf{p}_i[v].
\end{equation}

Using the average instead of the sum in the scoring function ensures that scores remain comparable across sequences with different numbers of remaining masked tokens.

\subsubsection{Position Selection Strategies}
Let $\mathcal{P}^{(j)}_t$ denote the set of candidate positions selected from $M^{(j)}_t$ for exploration. The positions are always ranked by confidence:
\begin{equation}
\begin{aligned}
\mathcal{P}^{(j)}_t &= \{i_1, i_2, \dots, i_{|\mathcal{P}^{(j)}_t|}\}, \\
\text{where } &\max_{v} \mathbf{p}^{(j)}_{t,i_1}[v] \geq \max_{v} \mathbf{p}^{(j)}_{t,i_2}[v] \geq \cdots
\end{aligned}
\end{equation}

\begin{figure*}[t]
    \centering
    \vskip 0.1in
    \includegraphics[width=0.99\textwidth]{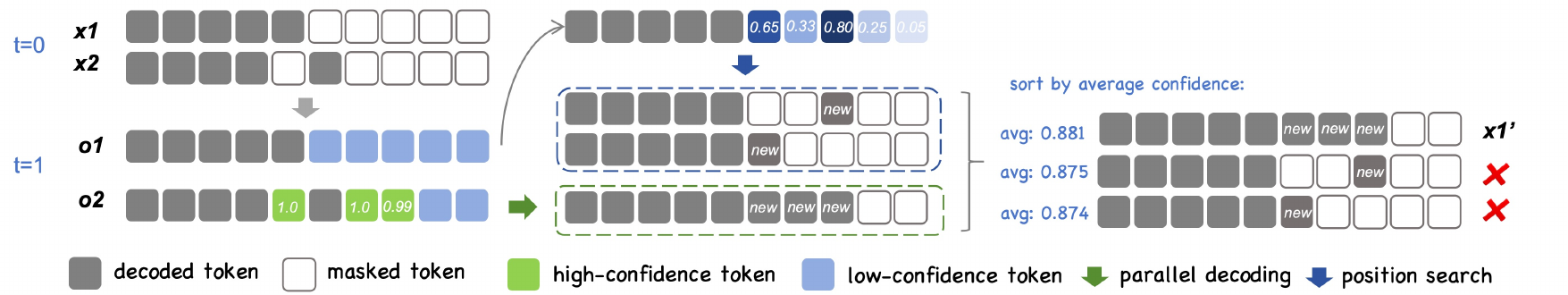}
    \vskip 0.1in
    \caption{Overview of \method. When t=0, there are two sequences in the beam. Based on the confidence of decoded tokens, one undergoes parallel decoding (green arrow), while the other undergoes position search (blue arrow). After reordering, since the sequence with the highest average confidence is obtained through parallel decoding, the beam size is reduced to 1.}
    \vskip 0.1in
    \label{fig:method}
\end{figure*}

\textbf{Case 1: Single-token Decoding ($n = 1$)} \\
% When only one token is unmasked per step, a candidate are generated by selecting only one token of $\mathbf{p}^{(j)}_t$. For each position $i \in \mathcal{P}^{(j)}_t$, unmasking only that position to generate a new condidate.

When only one token is unmasked per step, which is a boundary case of parallel decoding, candidates are generated by selecting only one token from $\mathbf{p}^{(j)}_t$, and the tokens selected by different candidates are mutually exclusive.

\textbf{Case 2: Parallel Decoding ($n > 1$)} \\
When $n$ tokens are unmasked per step, firslty, the minimum number of candidate tokens $M$ required to generate $K$ distinct combinations is determined:
\begin{equation}
M = \min\left\{ m : \binom{m}{n} \geq K, m \leq |M^{(j)}_t| \right\}
\end{equation}
If $|M^{(j)}_t| < M$, all masked positions are selected, so $M = |M^{(j)}_t|$. Then $\mathcal{P}^{(j)}_t$ contains the top-$M$ positions by confidence. To avoid the combinatorial explosion of generating all $\binom{M}{n}$ subsets, we efficiently select the top-$K$ $n$-element subsets from $\mathcal{P}^{(j)}_t$ by employing a best-first search strategy over the confidence-ordered positions, ensuring that only the most promising combinations are explored. These $K$ subsets are then used as candidate position sets for parallel unmasking.
\subsubsection{Beam Update}
All candidates from all beams are pooled, sorted by their scores, and the top $K$ form the new beam:
\begin{equation}
\mathcal{B}_{t-1} = \arg\max_{\mathcal{C} \subseteq \bigcup_{j=1}^{K} \mathcal{C}^{(j)}_t, |\mathcal{C}| = K} \sum_{(x,s) \in \mathcal{C}} s
\end{equation}

Table~\ref{tab:main_results} presents experimental results for single-token decoding and parallel decoding settings. In the single-token decoding setting, PBS consistently improves performance across multiple benchmarks. However, this improvement comes at a computational cost: computation scales linearly with beam size, resulting in proportionally slower decoding.

Enabling PBS with parallel decoding makes inference speed comparable to greedy decoding, yet leads to a noticeable drop in accuracy relative to the single-token-per-step setup. We hypothesize that this degradation stems from the confidence-agnostic nature of parallel decoding: by forcing the model to decode a fixed number of tokens at each step, it may be compelled to generate low-confidence tokens, thereby compromising output quality. We examine this hypothesis in more detail in Section~\ref{sec:analysis_of_decoding}.

To navigate this trade-off between decoding speed and output confidence, we introduce \method, a method that dynamically decides when to accelerate via parallel decoding and when to refine via confidence-based search.

\subsection{\method: Confidence-Switched Position Beam Search}
\method addresses this trade-off by dynamically adapting the decoding strategy according to prediction confidence, enabling the model to explore more when uncertain and decode faster when confident. Let $\mathcal{C}_t = \{(x^{(1)}_t, s^{(1)}_t), \dots, (x^{(N)}_t, s^{(N)}_t)\}$ denote the set of candidate sequences at step $t$, where $N_t = |\mathcal{C}_t|$ is the candidate size, each $x^{(j)}_t$ represents a candidate sequence, and $s^{(j)}_t$ is its cumulative score, with $j$ indexing candidates in the beam.

\subsubsection{Confidence-Guided Adaptive Decoding}
At each decoding step $t$, for each candidate sequence $x^{(j)}_t$ in the candidate set, we compute confidence scores for all masked positions:

\begin{equation}
c_{t,i} = \max_{v \in \mathcal{V}} \mathbf{p}_{t,i}[v]
\end{equation}

$x^{(j)}_t$ switches between two decoding modes based on the presence of high-confidence positions for each candidate sequence:

(a) \textbf{Parallel Decoding Mode}: When there exists at least one high-confidence position, \method directly decodes all such positions in parallel:

\begin{equation}
\mathcal{I}_t = \{i : c_{t,i} > \tau\}
\end{equation}

All tokens in $\mathcal{I}_t$ are unmasked simultaneously, where $\tau$ is a hyper-parameter. This mode is similar to the design of Fast-dLLM \cite{wu2025fastdllmtrainingfreeaccelerationdiffusion}, with the difference that when confidence falls below the threshold, \method will switch to beam search mode to explore other possible decoding paths.

(b) \textbf{Beam Search Mode}: When no masked position exceeds the confidence threshold ($c_{t,i} \leq \tau$ for all $i \in M^{(j)}_t$), \method falls back to position beam search with $k = 1$ token per step. The beam explores the top-$K$ most confident positions individually, trading speed for more thorough exploration.

\newcommand{\blueit}[1]{\textcolor{green!50!black}{\textit{#1}}}

\begin{table*}[t]
\centering
\vspace{0.1in}
\caption{Main Results. The upper section presents results on \textsc{LLaDA}-8B-Base, while the lower section shows results on \textsc{Dream}-7B-Base.}
\vspace{0.1in}
\label{tab:main_results}
\small
\setlength{\tabcolsep}{8pt}
\setlength{\extrarowheight}{2pt}
\resizebox{0.95\linewidth}{!}{%
\begin{tabular}{@{}lccccccc@{}}
\toprule
\multicolumn{1}{l}{Benchmark} & \multicolumn{2}{c}{HumanEval} & \multicolumn{2}{c}{MBPP} & \multicolumn{2}{c}{GSM8K} & Avg. \\
\cmidrule(lr){2-3} \cmidrule(lr){4-5} \cmidrule(lr){6-7} \cmidrule{8-8}
\multicolumn{1}{l}{Max Length} & 256 & 512 & 256 & 512 & 256 & 512 &  \\
\midrule

\textbf{Greedy} & 32.3 & 32.9 & 40.8 & 39.2 & 70.4 & 70.9 & 47.8 \\
\addlinespace[0.3em]

\textbf{Greedy (Adaptive Parallel)} & 32.3 & 32.9 & 40.8 & 39.2 & 70.4 & 71.0 & 47.8 \\
\textit{SpeedUp} & \blueit{×2.25} & \blueit{×2.79} & \blueit{×2.06} & \blueit{×2.42} & \blueit{×1.69} & \blueit{×1.92} & \blueit{×2.19} \\
\addlinespace[0.3em]

\textbf{PBS (Single Token)} & 34.2 & 34.8 & 41.4 & 39.2 & 71.6 & 72.1 & 48.9 \\
\textit{SpeedUp} & \blueit{×0.48} & \blueit{×0.48} & \blueit{×0.48} & \blueit{×0.48} & \blueit{×0.47} & \blueit{×0.48} & \blueit{×0.48} \\
\addlinespace[0.3em]

\textbf{PBS (Parallel)} & 31.1 & 29.3 & 35.2 & 36.1 & 69.5 & 68.1 & 44.9 \\
\textit{SpeedUp} & \blueit{×0.98} & \blueit{×0.98} & \blueit{×0.97} & \blueit{×0.97} & \blueit{×0.98} & \blueit{×0.99} & \blueit{×0.98} \\
\addlinespace[0.3em]

\rowcolor{blue!10}
\textbf{\method} & 32.9(+0.6) & 39.0(+6.1) & 40.8 & 39.4(+0.2) & 71.3(+0.9) & 71.5(+0.6) & 49.2(+1.4) \\
\rowcolor{blue!10}
\textit{SpeedUp} & \blueit{×1.63} & \blueit{×2.16} & \blueit{×1.49} & \blueit{×1.85} & \blueit{×1.18} & \blueit{×1.43} & \blueit{×1.62} \\
\midrule

\textbf{Greedy} & 50.0 & 53.7 & 53.4 & 55.4 & 73.7 & 74.5 & 60.1 \\
\addlinespace[0.3em]

\textbf{Greedy (Adaptive Parallel)} & 50.0 & 51.8 & 53.8 & 55.6 & 73.2 & 74.6 & 59.8 \\
\textit{SpeedUp} & \blueit{×1.63} & \blueit{×1.60} & \blueit{×2.20} & \blueit{×2.86} & \blueit{×2.19} & \blueit{×1.83} & \blueit{×2.05} \\
\addlinespace[0.3em]

\textbf{PBS (Single Token)} & 57.3 & 58.1 & 54.0 & 56.9 & 75.2 & 76.4 & 63.0 \\
\textit{SpeedUp} & \blueit{×0.57} & \blueit{×0.52} & \blueit{×0.50} & \blueit{×0.52} & \blueit{×0.57} & \blueit{×0.56} & \blueit{×0.54} \\
\addlinespace[0.3em]

\textbf{PBS (Parallel)} & 53.7 & 54.3 & 53.6 & 54.0 & 73.5 & 75.1 & 60.7 \\
\textit{SpeedUp} & \blueit{×1.02} & \blueit{×0.98} & \blueit{×1.01} & \blueit{×0.96} & \blueit{×1.03} & \blueit{×1.01} & \blueit{×1.00} \\
\addlinespace[0.3em]

\rowcolor{blue!10}
\textbf{\method} & 55.5(+5.5) & 55.1(+1.4) & 57.0(+3.6) & 56.2(+0.8) & 74.7(+1.0) & 75.7(+1.2) & 62.4(+2.3) \\
\rowcolor{blue!10}
\textit{SpeedUp} & \blueit{×1.13} & \blueit{×1.07} & \blueit{×1.87} & \blueit{×2.47} & \blueit{×0.95} & \blueit{×1.07} & \blueit{×1.43} \\
\bottomrule
\end{tabular}}
\end{table*}

% \subsubsection{Dynamic Beam Width Adjustment}
% soar further optimizes efficiency by adapting the beam width $B_t$ after each step:

% \begin{equation}
% B_t = \begin{cases}
% 1 & \text{if } \mathcal{I}_t \neq \emptyset \quad \text{(After Parallel Decoding)} \\
% K & \text{otherwise} \quad \text{(After Beam Search)}
% \end{cases}
% \end{equation}

% This adaptive strategy ensures that when the model is highly confident, computational resources are focused on a single reliable path; when uncertain, multiple decoding possibilities are explored to avoid local optima. Algorithm \ref{alg:soar} outlines soar's adaptive inference process.

% The time complexity of soar is $O(T \cdot \bar{B} \cdot L)$, where $T$ is the number of decoding steps, $L$ is the sequence length, and $\bar{B}$ is the average beam width. In practice, $\bar{B} < K$ due to frequent parallel decoding, and the reduced number of steps $T$ further accelerates inference.

\subsubsection{Dynamic Candidate Size Adjustment}
Sequences in the candidate set generate new candidate sequences, which are then sorted in descending order based on their average decoding confidence. After sorting, only the top $N_t$ sequences are retained, where $N_t$ denotes the size of the candidate set at step $t$.

To further optimize efficiency, \textsc{SOAR} dynamically adjusts $N_t$ after each decoding step based on the origin of the new candidate with the highest score. The decision rule is as follows: if the highest score new candidate originates from the parallel decoding mode, the model is considered sufficiently confident and $N_t$ is reduced to 1 for the next step; otherwise, if the best new candidate comes from the PBS mode, $N_t$ is reset to the preset size $K$, which corresponds to the maximum beam width used in the PBS mode. Formally,

\begin{equation}
N_t = \begin{cases}
1 & \text{if best new candidate from parallel decoding} \\
K & \text{if best new candidate from PBS}
\end{cases}
\end{equation}

This adaptive strategy ensures that when the model exhibits high confidence, computational resources are concentrated on the most promising path; when uncertainty arises, multiple decoding hypotheses are retained to avoid local optima. 

% The complete inference process is outlined in Algorithm~\ref{alg:soar}.

The time complexity of \textsc{SOAR} is $O(T \cdot \bar{N})$, where $T$ is the number of decoding steps, and $\bar{N}$ is the average candidate size. In practice, because a significant portion of the model's tokens have confidence above the threshold, $\bar{N}$ remains well below $K$, and the overall step count $T$ is also reduced. 
% This efficiency can be formally characterized by two extreme cases:
% \begin{proposition}[Efficiency Bounds]
% \textsc{SOAR} achieves its upper efficiency bound when at every step there exists at least one position with confidence above the threshold, effectively reducing to confidence-based parallel decoding with time complexity $O(T' \cdot 1)$, where $T'$ is the reduced number of steps. Conversely, it reaches its lower efficiency bound when at any step no position exceeds the confidence threshold, reverting to position beam search with complexity $O(T_{\text{greedy}} \cdot K)$, where $T_{\text{greedy}}$ is the number of steps in greedy decoding.
% \end{proposition}

\textsc{SOAR} thus provides a principled way to balance the exploration-exploitation tradeoff in DLM decoding, adapting to the model's uncertainty at each step to optimize both quality and efficiency. Appendix~\ref{app:illustration} presents a visual analysis of the algorithm's behavior.

\section{Experiments}

\subsection{Experiment Setup}

% To ensure reproducibility, all experiments are conducted with a temperature of $0$, using softmax probabilities as confidence scores.

We conduct experiments on two representative diffusion language models: \textsc{LLaDA}-8B \cite{nie2025largelanguagediffusionmodels} and \textsc{Dream}-7B \cite{ye2025dream7bdiffusionlarge}, both using the Base version.  Unless otherwise specified, we use the softmax probabilities as confidence scores. The confidence threshold \(\tau\) is set to \(0.95\) for \textsc{LLaDA}-8B and \(0.90\) for \textsc{Dream}-7B, and the maximum beam size \(K\) is set to \(2\). The maximum sequence length is set to $256$ and $512$, respectively, to validate the robustness of the results.

To comprehensively assess the effectiveness of our approach, experiments are conducted across three benchmarks covering code generation and mathematical reasoning. For mathematical reasoning, we employ GSM8K \cite{cobbe2021trainingverifierssolvemath}, a dataset of grade-school math word problems. For code generation, we evaluate on MBPP \cite{austin2021programsynthesislargelanguage}, which contains entry-level Python programming tasks, and HumanEval \cite{chen2021evaluatinglargelanguagemodels}, a collection of handwritten programming challenges designed for synthesis evaluation.

\begin{figure}[t]
    \centering
    \includegraphics[width=0.95\linewidth]{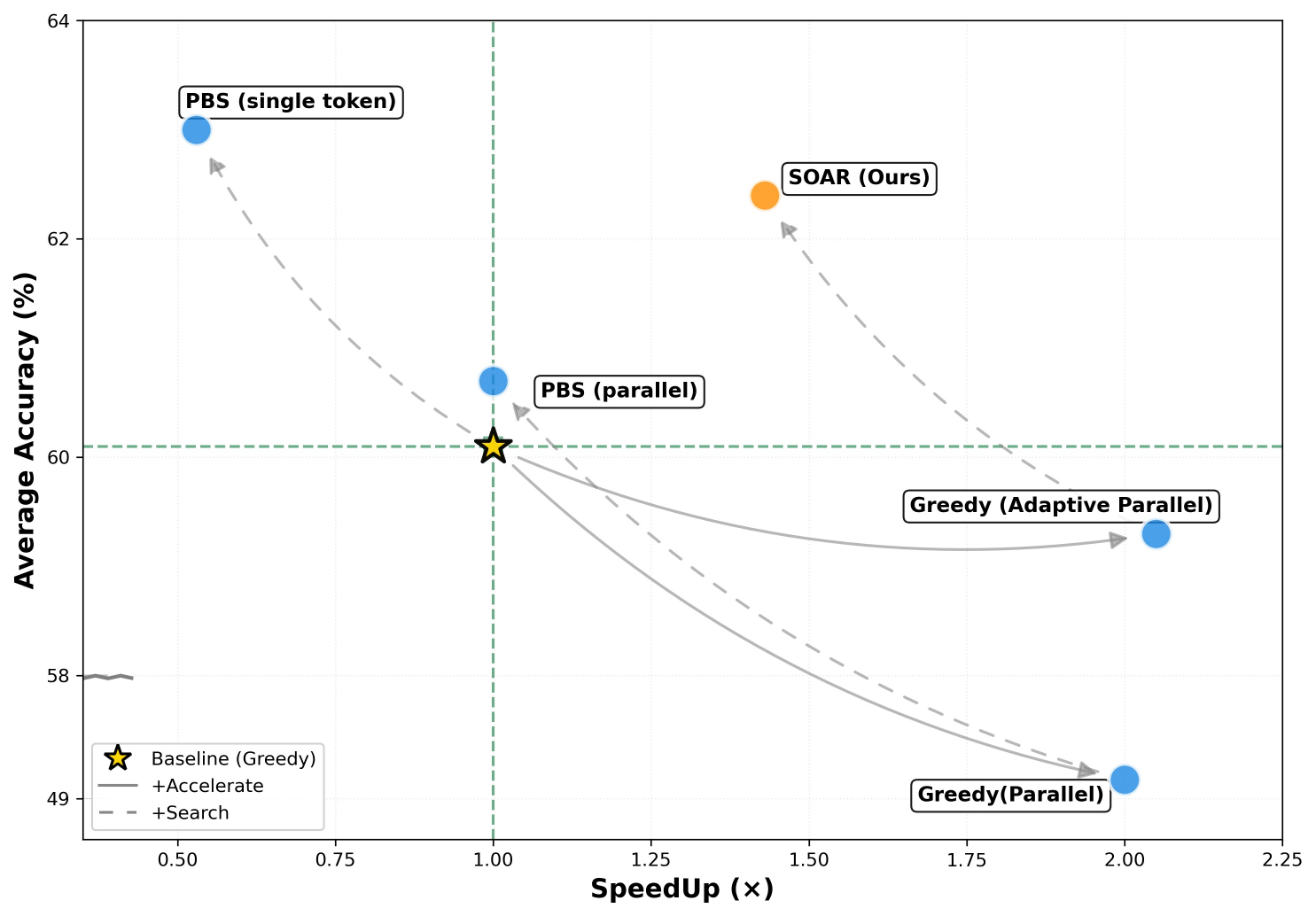}
    \vskip 0.1in
    \caption{Pareto frontier on \textsc{Dream}-7B-Base. Solid arrows indicate adding parallel decoding (acceleration), and dashed arrows indicate adding position beam search. This plot is generated using the average accuracy and average speedup from Table~\ref{tab:main_results}. For better visual presentation, the Y-axis range 49–58 is compressed.}
    \vskip 0.1in
    \label{fig:main_results}
\end{figure}

We adhere to standard few-shot evaluation protocols for each benchmark: 0-shot for HumanEval, 3-shot for MBPP and 4-shot for GSM8K. Accuracy is used as the evaluation metric for mathematical reasoning, while pass@1 is adopted for code generation benchmarks. All experiments are conducted based on open-source lm-evaluation-harness \cite{eval-harness} with a single NVIDIA A100 80GB GPU. The total inference time of the benchmark is used to calculate the SpeedUp ratio compared with standard greedy decoding.

\begin{figure*}[t]
    \centering
    \includegraphics[width=0.98\textwidth]{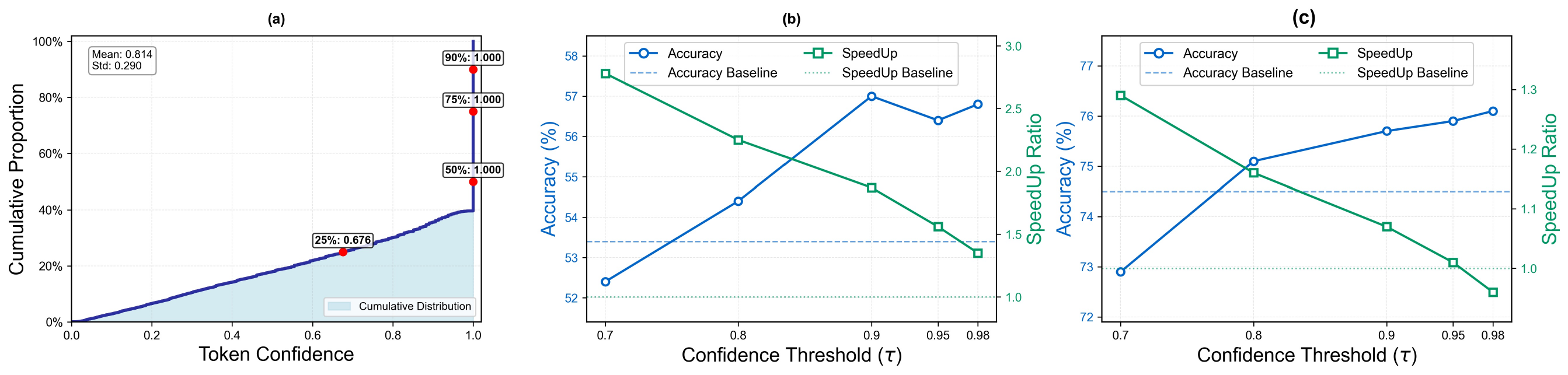}
    \vskip 0.1in
    \caption{The threshold study on \textsc{Dream-7B}:  (a) Cumulative distribution of token confidence scores on GSM8K; (b) Trade-off between accuracy and SpeedUp under varying confidence thresholds on MBPP; (c) Trade-off between accuracy and SpeedUp under varying confidence thresholds on GSM8K.}
    \vskip 0.1in
    \label{fig:confidence_ablation}
\end{figure*}

\subsection{Main Results on Accuracy and Speed}

Table~\ref{tab:main_results} summarizes the performance of different decoding methods across both models and three benchmarks. In addition, we plot the variation of average accuracy and SpeedUp among these decoding methods in Figure~\ref{fig:main_results} for better illustration. In most settings, PBS consistently improves accuracy over greedy decoding, with \textsc{Dream}-7B achieving gains of up to $+7.3\%$ on HumanEval. However, this quality improvement comes at the cost of nearly doubled inference latency (average SpeedUp $\times 0.54$), as exploring multiple decoding paths increases computation.

PBS with parallel token decoding ($n=2$) restores decoding speed to the same level as greedy decoding, yet the average accuracy also decreases significantly compared to PBS with single token. We attribute this to the fact that forcing a fixed number of tokens to be decoded in parallel can undermine the confidence-based selection that underlies PBS's effectiveness. In contrast, \method switches between search and parallel decoding based on confidence, avoiding excessive search that affects decoding speed while also preventing mandatory parallel decoding from harming global confidence, thereby achieving a balance between speed and quality improvement.

\subsection{Ablation Study}

\textbf{Confidence Threshold.} We examine how the confidence threshold $\tau$ affects decoding quality and speed. Figure.~\ref{fig:confidence_ablation}(a) plots the cumulative distribution of token confidence scores on GSM8K under standard decoding, showing a long tail—about 50\% of tokens have probability $>0.95$. We vary $\tau \in [0.8, 0.98]$ and evaluate accuracy and SpeedUp with coding (MBPP) and math (GSM8K) tasks on \textsc{Dream}-7B. As shown in Figure.~\ref{fig:confidence_ablation}(b)-(c), raising $\tau$ from $0.8$ to $0.98$ steadily improves accuracy but reduces decoding speed. Notably, for $\tau > 0.8$, \method consistently beats standard decoding in accuracy while matching or exceeding its speed, demonstrating robustness. We conducted the same analysis on \textsc{LLaDA}-8B, and ultimately set $\tau = 0.95$ for \textsc{LLaDA}-8B and $\tau = 0.9$ for \textsc{Dream}-7B.

\begin{figure}[t]
    \centering
    \includegraphics[width=0.66\linewidth]{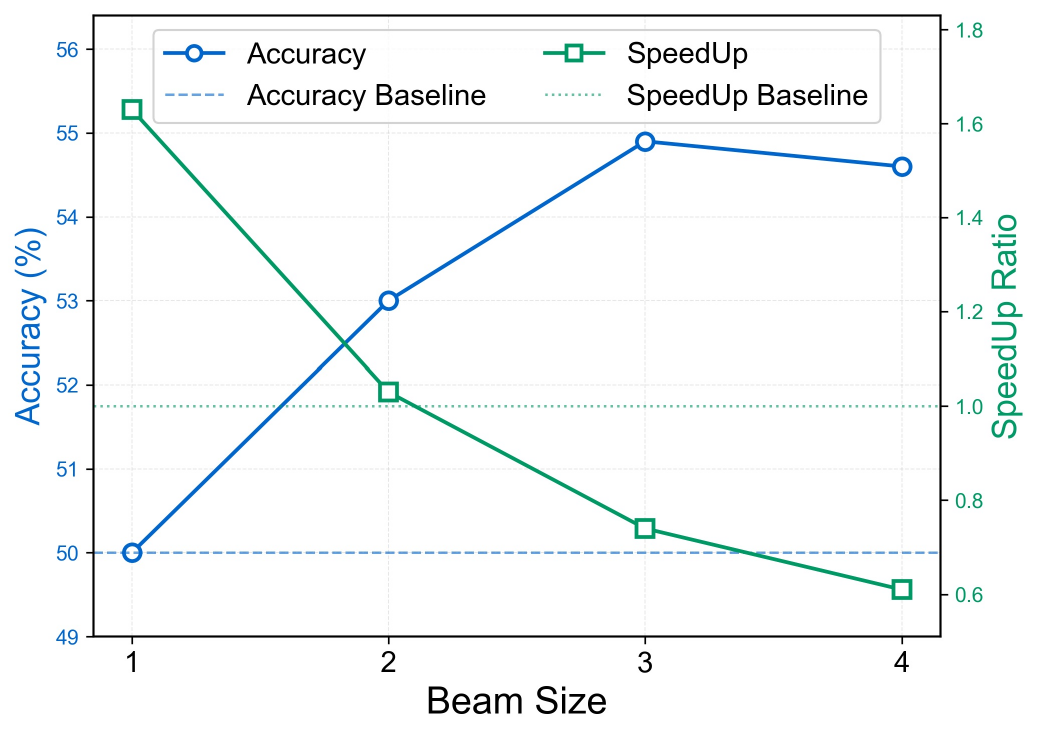}
    \vskip 0.1in
    \caption{The beam size study on \textsc{Dream}-7B: Trade-off between accuracy and SpeedUp under increasing beam size on HumanEval.}
    \vskip 0.1in
    \label{fig:beamsize_ablation}
\end{figure}

\textbf{Beam Size.} We further investigate whether expanding the beam size can yield additional performance gains. Figure~\ref{fig:beamsize_ablation} illustrates the trade-off between generation quality and inference speed under different beam sizes (ranging from 1 to 4) on the HumanEval benchmark using the \textsc{Dream}-7B model. Setting beam size to 1 corresponds to using only confidence-based parallel decoding. While increasing the beam size may offer modest improvements in accuracy, the computational overhead grows linearly with beam width, resulting in significant slowdowns in inference speed. Therefore, we select a beam size of 2 as the default setting to balance quality and efficiency in our experiments.

\subsection{Analysis of Confidence and AR-ness}
\label{sec:analysis_of_decoding}
\begin{figure*}[t]
    \centering
    \includegraphics[width=0.99\textwidth]{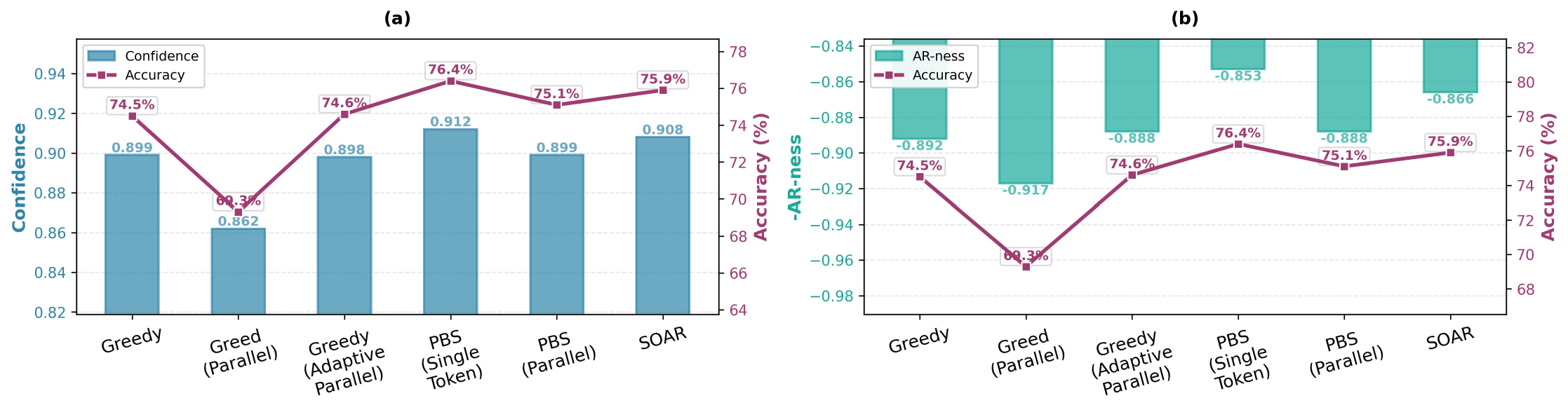}
    \vskip 0.1in
    \caption{Left: Accuracy with average confidence. Right: Accuracy with negative Global AR-ness}
    \vskip 0.1in
    \label{fig:confidence_analysis}
\end{figure*}

\textbf{Confidence of Decoding Sequence.} 
To test the hypothesis proposed in Section~\ref{sec:introduction}—\textit{Can we improve decoding quality by exploring alternative unmasking sequences with higher confidence beyond greedy selection?}—we analyze the average confidence of decoded sequences.

\begin{definition}
  \label{def:average_confidence}
  Average Confidence: For each sample, we extract the reasoning tokens that precede the answer keyword (``answer'' token in GSM8K). Each token's confidence score is defined as the model's maximum softmax probability. The sample-level average confidence is computed as the mean of these token-wise scores. We then average across all samples for each decoding method to obtain the method-level average confidence.
\end{definition}

Figure~\ref{fig:confidence_analysis}(a) illustrates the relationship between average confidence and accuracy for various decoding methods on the GSM8K benchmark using \textsc{Dream}-7B. Here, ``Parallel'' refers to decoding exactly two tokens per step, ``Adaptive Parallel'' decides whether to perform parallel decoding based on confidence thresholds, and the maximum sequence length is 512.

Our results show a positive correlation between accuracy and average confidence. PBS achieves an average accuracy 1.9\% higher than Greedy Search, confirming our hypothesis. In contrast, Parallel decoding forces the generation of two tokens per step regardless of confidence, which significantly lowers average confidence and leads to a notable accuracy drop. When Parallel decoding is combined with PBS, the confidence gains from PBS are negated by the confidence loss from Parallel decoding, resulting in a 1.3\% accuracy reduction compared to PBS alone. \method, however, dynamically switches between Parallel decoding and PBS based on confidence, preserving higher average confidence and maintaining competitive accuracy.

\textbf{AR-ness of Decoding Sequence.}
We further investigate why \method yields better decoding sequences using another metric: the Global AR-ness proposed by  \cite{gong2025diffucoderunderstandingimprovingmasked}, which quantifies how much the unmasking schedule of a diffusion model resembles an autoregressive pattern, specifically, whether it follows a ``left-first'' pattern.

\begin{definition}
  \label{def:global_arness}
  Global AR-ness: At each decoding step $t$, we examine whether the predicted token lies within the first $k$ masked positions. The Global AR-ness@$k$ is defined as the average ratio of such steps across the entire decoding process, measuring the tendency to always unmask the earliest remaining token and thus capturing a left-to-right filling strategy. This ratio increases with $k$, as the criterion becomes easier to satisfy when more early positions are allowed. \textit{A higher value indicates that the generation is more autoregressive.}
\end{definition}

We set $k=5$ and compute the mean Global AR-ness across all samples. Figure~\ref{fig:confidence_analysis}(b) illustrates the relationship between the negative Global AR-ness and accuracy, revealing a \textbf{negative} correlation: methods with lower Global AR-ness (i.e., less left-to-right bias) tend to achieve higher accuracy. We hypothesize that this may be because \method can mitigate the “entropy sink” issue in DLMs. Specifically, the model is inherently biased toward tokens immediately to the right of the given prefix, as these positions receive stronger positional signals and closer context, leading to disproportionately high confidence. This bias may limit the DLM's ability to explore potentially better decoding paths. Although \method is not explicitly designed to address this problem, by enabling position beam search, it can alleviate the issue.

\subsection{Analysis of the Robustness of \method}
\label{sec:analysis_of_metric}
\begin{table}[t]
\centering
\vspace{0.1in}
\caption{Results of other unmask metrics on HumanEval with \textsc{Dream}-7B.}
\vspace{0.1in}
\label{tab:metric_ablation}

\resizebox{0.95\linewidth}{!}{%
\begin{tabular}{lccc}
\toprule
\textbf{Metric} & \textbf{Method} & \textbf{Length=256} & \textbf{Length=512} \\
\midrule
Margin & Greedy & 47.0 & 49.4 \\
Margin & \method & 51.2 & 51.2 \\
 & \textit{SpeedUp} & \blueit{×1.04} & \blueit{×1.07} \\
\hline
NegEntropy & Greedy & 51.2 & 56.1 \\
NegEntropy & \method & 53.0 & 57.9 \\
 & \textit{SpeedUp} & \blueit{×1.07} & \blueit{×1.14} \\
\bottomrule
\end{tabular}}
\end{table}

\textbf{Unmask Metric.} Although softmax probability is the mainstream metric for selecting unmask tokens, several alternative metrics are also viable options: (1) the margin between the top‑1 and top‑2 token probabilities and (2) the negative entropy of the probability distribution. Formally, given the token probability distribution $\mathbf{p} \in \mathbb{R}^V$, the two metrics are defined as
\[
\text{Margin}(\mathbf{p}) = p_{(1)} - p_{(2)},
\]
\[
\text{NegEntropy}(\mathbf{p}) = \sum_{i=1}^{V} p_i \log p_i,
\]
where $p_{(1)}$ and $p_{(2)}$ denote the largest and second‑largest probabilities in $\mathbf{p}$, respectively.

Table~\ref{tab:metric_ablation} presents the results on HumanEval under two sequence length settings.  We use $\tau=0.9$ for margin‑based metric and $\tau=-0.1$ for negative‑entropy‑based metric. For both, \method consistently improves accuracy over the corresponding greedy decoding baseline while maintaining a faster decoding speed.

% \subsection{Robustness to Variable-Length Decoding}

\begin{table}[htbp]
    \centering
    \caption{Results on HumanEval-Infilling with \textsc{DreamOn} under different initial sequence lengths.}
    \resizebox{0.75\linewidth}{!}{%
    \begin{tabular}{lcccc}
        \toprule
        & \multicolumn{4}{c}{\textbf{Initial Mask Length}} \\
        \cmidrule(lr){2-5}
        \textbf{Method} & \textbf{8} & \textbf{16} & \textbf{32} & \textbf{64} \\
        \midrule
        Greedy & 58.4 & 57.4 & 58.0 & 58.0 \\
        \method & 63.0 & 66.1 & 67.2 & 67.7 \\
        \textit{SpeedUp} & \blueit{x1.02} & \blueit{x1.02} & \blueit{x1.03} & \blueit{x1.04} \\
        \bottomrule
    \end{tabular}
    \label{tab:var_length}
    }
\end{table}

\textbf{Results with Variable-Length Decoding Method}. To assess the robustness of \method when combined with variable-length decoding strategies, we evaluate it on \textsc{DreamOn} \cite{Dreamon2025}---a model post-trained upon \textsc{Dream}-7B that dynamically adjusts sequence length during decoding by predicting \textit{expand} and \textit{delete} tokens. Specifically, when an \textit{expand} token is generated, it is replaced with two \textit{mask} tokens, thereby extending the sequence; conversely, a \textit{delete} token triggers the removal of the corresponding token from the output sequence.

Following the DreamOn experimental setup, we conduct evaluations on the HumanEval-Infilling benchmark \cite{bavarian2022efficienttraininglanguagemodels}. We set the maximum decoding length to 64, employ negative entropy ($\tau = -0.1$) as the confidence metric, and use exact match as the evaluation criterion.

Table~\ref{tab:var_length} compares the performance of standard greedy decoding and \method across different initial sequence lengths. The results demonstrate that \method, as a training-free method, consistently maintains or improves performance when integrated with DreamOn’s variable-length decoding mechanism. This confirms \method’s flexibility and robustness, highlighting its potential to be effectively combined with other carefully designed decoding strategies without requiring architectural modifications or additional training.

% \textbf{Results with RL Model}. doing
% benchmark: GSM8K, model: d1

\section{Conclusion}

We introduce \methodfull, an adaptive inference framework that dynamically balances generation quality and speed for DLMs. By allowing the model to widen its search when uncertain and accelerate when confident, \method consistently improves generation quality across coding and mathematical reasoning tasks while maintaining competitive inference speed. Ablation studies further validate the robustness of \method under varying confidence thresholds, beam sizes, and unmask metrics. Moreover, \method generalizes effectively across different base models and sequence lengths, and integrates seamlessly with variable-length decoding strategies like \textsc{DREAMON}. As a training-free method, \method offers a practical and effective solution to the quality-speed trade-off in DLM inference.

% \section*{Impact Statement}

% This paper presents work whose goal is to advance the field of Machine
% Learning. There are many potential societal consequences of our work, none which we feel must be specifically highlighted here.

\bibliography{main}
\bibliographystyle{icml2026}

%%%%%%%%%%%%%%%%%%%%%%%%%%%%%%%%%%%%%%%%%%%%%%%%%%%%%%%%%%%%%%%%%%%%%%%%%%%%%%%
%%%%%%%%%%%%%%%%%%%%%%%%%%%%%%%%%%%%%%%%%%%%%%%%%%%%%%%%%%%%%%%%%%%%%%%%%%%%%%%
% APPENDIX
%%%%%%%%%%%%%%%%%%%%%%%%%%%%%%%%%%%%%%%%%%%%%%%%%%%%%%%%%%%%%%%%%%%%%%%%%%%%%%%
%%%%%%%%%%%%%%%%%%%%%%%%%%%%%%%%%%%%%%%%%%%%%%%%%%%%%%%%%%%%%%%%%%%%%%%%%%%%%%%
\newpage
\appendix
\onecolumn

\section{Visualization of \textsc{SOAR}}
\label{app:illustration}

\begin{figure*}[h]
    \centering
    \includegraphics[width=0.75\textwidth]{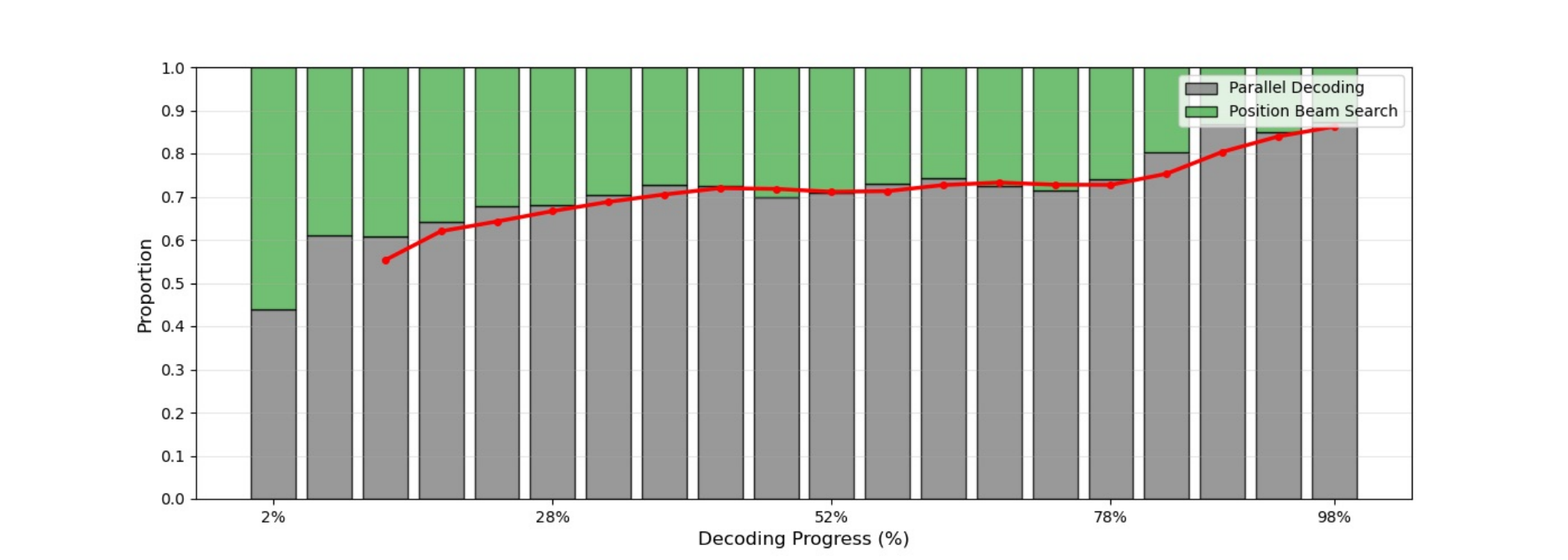}
    \caption{Variation of Decoding Modes Across the Decoding Process. The x-axis represents the progress of decoding (quantized into 20 bins), while the y-axis indicates the proportion of tokens decoded by each mode across all samples. The stacked bars are partitioned into gray (Parallel Decoding) and green (PBS) segments, reflecting their relative usage at each stage. A red trend line highlights the overall transition pattern between the two modes.}
    \label{fig:decoding_process_with_parallel_ratio}
\end{figure*}

\textbf{Increasingly Parallel Decoding.} 
We analyzed the usage patterns of Beam Search and Parallel Decoding modes across decoding steps on the Dream-7B-Base model using GSM8K. Consistent with the processing in Subsection~\ref{sec:analysis_of_decoding}, for all sample decoding sequences, we retain only the valid portion, i.e., the tokens preceding the keyword ``answer''. Since different samples have varying effective decoding lengths \( S_{\text{sample}} \), we record the ratio of each token's decoding step to its effective sequence length as a measure of the Decoding Process (\%). Figure~\ref{fig:decoding_process_with_parallel_ratio} illustrates how the two decoding modes evolve as decoding progresses. In the early stages, \textsc{SOAR} allocates a larger proportion of steps (55\%) to PBS to explore more promising sequences. As decoding advances, the model eventually favors parallel decoding in 85\% of the cases. This trend demonstrates the model's transition from initial uncertainty toward greater confidence. Such behavior is expected, as the progressive reduction in remaining masked tokens increasingly constrains the conditional distribution over subsequent decoding steps, resulting in a more concentrated distribution and higher model confidence.
Throughout the inference process, \textsc{SOAR} dynamically switches between the two decoding modes based on confidence scores in an adaptive manner.

\begin{figure*}[htbp]
    \centering
    \includegraphics[width=0.75\textwidth]{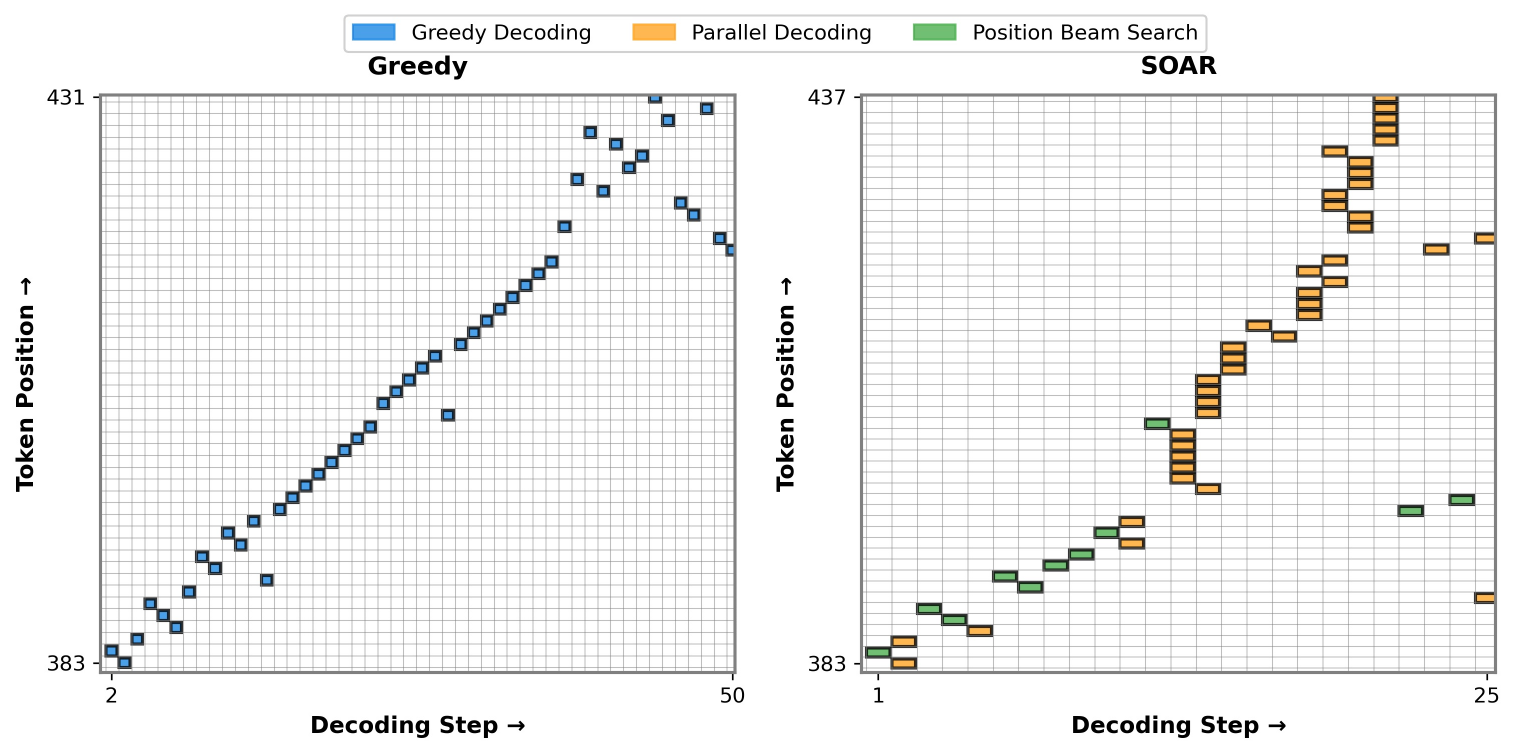}
    \vspace{-0.5em}
    \caption{Decoding Path Comparison. The left figure shows Greedy Decoding, and the right figure shows Our method. The X-axis represents steps, and the Y-axis represents positions.}
    \label{fig:decoding_paths_comparision}
    
\end{figure*}

\textbf{Decoding Path Illustration}. Figure~\ref{fig:decoding_paths_comparision} illustrates the decoding trajectories of a GSM8K sample using Greedy and \textsc{SOAR} decoding, based on the Dream-7B-Base model. The Greedy decoder exhibits an AR-style decoding path, while \textsc{SOAR}, through parallel decoding, decodes nearly the same number of positions in fewer steps. Moreover, by employing PBS, \textsc{SOAR} retains earlier positions as masked and defers their decoding to later steps, aligning with the analysis in Subsection~\ref{sec:analysis_of_decoding}. For tokens that appear early but lack sufficient model confidence, decoding them later with richer contextual information yields an overall higher-confidence sequence compared to Greedy decoding.

\section{Pseudocode of SOAR}
\label{app:algo}
Algorithm~\ref{alg:soar} presents the complete decoding procedure of \method, which dynamically switches between parallel decoding and position beam search based on token-level confidence.

\begin{algorithm}[t]
\caption{\methodfull}
\label{alg:soar}
\begin{algorithmic}[1]
\REQUIRE Prompt $x_{\text{prompt}}$, DLM $f_\theta$, max length $L$, confidence threshold $\tau$, max beam width $B_{\text{max}}$
\ENSURE Generated sequence $x_0$
\STATE Initialize $x_T$ with $x_{\text{prompt}}$ followed by $L$ [MASK] tokens
\STATE Initialize beam $\mathcal{B}_T = \{(x_T, 0)\}$
\STATE Set candidate size $N_T = 1$ \COMMENT{Initial candidate size}
\FOR{$t = T, T-1, \dots, 1$}
    \STATE $\mathcal{C}_t \leftarrow \emptyset$ \COMMENT{Candidate set for step $t$}
    
    \FOR{each $(x_t^{(j)}, s_t^{(j)}) \in \mathcal{B}_t$}
        \STATE Compute token distributions $\mathbf{p}_t^{(j)} = f_\theta(x_t^{(j)})$
        \STATE Compute confidence scores $c_{t,i}^{(j)} = \max_v \mathbf{p}_{t,i}^{(j)}[v]$ for all positions $i$
        \STATE Identify masked positions $M_t^{(j)} = \{i: x_{t,i}^{(j)} = \text{[MASK]}\}$
        \STATE Identify high-confidence positions $H_t^{(j)} = \{i \in M_t^{(j)}: c_{t,i}^{(j)} > \tau\}$
        
        \IF{$H_t^{(j)} \neq \emptyset$}
            \STATE \COMMENT{--- Parallel Decoding Mode ---}
            \STATE $\mathcal{I} \leftarrow H_t^{(j)}$ \COMMENT{Unmask all high-confidence positions}
            \STATE $x_{t-1} \leftarrow \text{Unmask}(x_t^{(j)}, \mathcal{I}, \mathbf{p}_t^{(j)})$
            \STATE $\Delta s \leftarrow \frac{1}{|\mathcal{I}|}\sum_{i \in \mathcal{I}} c_{t,i}^{(j)}$
            \STATE Add $(x_{t-1}, s_t^{(j)} + \Delta s, \textsc{Parallel})$ to $\mathcal{C}_t$
        \ELSE
            \STATE \COMMENT{--- Position Beam Search Mode ---}
            \STATE Sort masked positions by confidence: $i_1, i_2, \dots$ where $c_{t,i_1}^{(j)} \geq c_{t,i_2}^{(j)} \geq \cdots$
            \STATE $k \leftarrow \min(B_{\text{max}}, |M_t^{(j)}|)$
            \FOR{$\ell = 1, \dots, k$}
                \STATE $x_{t-1}^{(\ell)} \leftarrow \text{Unmask}(x_t^{(j)}, \{i_\ell\}, \mathbf{p}_t^{(j)})$
                \STATE Add $(x_{t-1}^{(\ell)}, s_t^{(j)} + c_{t,i_\ell}^{(j)}, \textsc{BeamSearch})$ to $\mathcal{C}_t$
            \ENDFOR
        \ENDIF
    \ENDFOR
    
    \STATE \COMMENT{--- Candidate Pruning and Beam Update ---}
    \STATE Sort $\mathcal{C}_t$ by average cumulative score in descending order
    \STATE Let $(x^*, s^*, \text{mode}^*)$ be the top candidate in $\mathcal{C}_t$
    
    \IF{$\text{mode}^* = \textsc{Parallel}$}
        \STATE \COMMENT{High confidence: focus on the best path}
        \STATE Set candidate size $N_t \leftarrow 1$
    \ELSE
        \STATE \COMMENT{Uncertainty detected: maintain multiple hypotheses}
        \STATE Set candidate size $N_t \leftarrow \min(B_{\text{max}}, |\mathcal{C}_t|)$
    \ENDIF
    
    \STATE $\mathcal{B}_{t-1} \leftarrow$ top $N_t$ candidates from $\mathcal{C}_t$
\ENDFOR
\STATE \textbf{return} $x_0$ from the best candidate in $\mathcal{B}_0$
\end{algorithmic}
\end{algorithm}

\end{document}